\documentclass[lettersize,journal]{IEEEtran}
\usepackage{amsmath,amsfonts}
\usepackage{algorithm}
\usepackage{array}
\usepackage[caption=false,font=normalsize,labelfont=sf,textfont=sf]{subfig}
\usepackage{textcomp}
\usepackage{stfloats}
\usepackage{url}
\usepackage{verbatim}
\usepackage{graphicx}
\usepackage{cite}
\usepackage{verbatim}
\usepackage{amsmath}
\usepackage{color,array}
\usepackage{multirow}
\usepackage{diagbox}
\usepackage{amssymb}
\usepackage{enumitem} 
\setlist[itemize]{noitemsep}
\usepackage{algpseudocode}
\usepackage{hyperref}

\newcommand{\vect}[1]{\boldsymbol{#1}}

\newcommand{\blue}[1]{\textcolor{blue}{#1}}

\hypersetup{
    colorlinks=true,
    citecolor=blue,      % Citation color (brackets and numbers)
    linkcolor=blue,      % Internal document links (e.g., table of contents)
    urlcolor=blue        % External URLs
}

\begin{document}

\title{Semantic Segmentation by Semantic Proportions}

\author{Halil Ibrahim Aysel, Xiaohao Cai, Adam Prugel-Bennett
%\author{IEEE Publication Technology,~\IEEEmembership{Staff,~IEEE,}
        % <-this % stops a space
\thanks{The authors are with the School of Electronics and Computer Science, University of Southampton, UK. (Email: hia1v20@soton.ac.uk; x.cai@soton.ac.uk; and apb@ecs.soton.ac.uk)}% <-this % stops a space
%\thanks{Manuscript received April 19, 2021; revised August 16, 2021.}
}

% The paper headers
\markboth{Journal of \LaTeX\ Class Files,~Vol.~xx, No.~x, xxxx}%
{Shell \MakeLowercase{\textit{et al.}}: A Sample Article Using IEEEtran.cls for IEEE Journals}

%\IEEEpubid{0000--0000/00\$00.00~\copyright~2021 IEEE}
% Remember, if you use this you must call \IEEEpubidadjcol in the second
% column for its text to clear the IEEEpubid mark.

\maketitle

\begin{abstract}
Semantic segmentation is a critical task in computer vision aiming to identify and classify individual pixels in an image, with numerous applications in for example autonomous driving and medical image analysis. However, semantic segmentation can be highly challenging particularly due to the need for large amounts of annotated data. Annotating images is a time-consuming and costly process, often requiring expert knowledge and significant effort; moreover, saving the annotated images could dramatically increase the storage space. In this paper, we propose a novel approach for semantic segmentation, requiring the rough information of individual semantic class proportions, shortened as {\it semantic proportions},  rather than the necessity of ground-truth segmentation maps. This greatly simplifies the data annotation process and thus will significantly reduce the annotation time, cost and storage space, opening up new possibilities for semantic segmentation tasks where obtaining the full ground-truth segmentation maps may not be feasible or practical. Our proposed method of utilising semantic proportions can {\it (i)} further be utilised as a booster in the presence of ground-truth segmentation maps to gain performance without extra data and model complexity, and {\it (ii)} also be seen as a parameter-free plug-and-play  module, which can be attached to existing deep neural networks designed for semantic segmentation. Extensive experimental results demonstrate the good performance of our method compared to benchmark methods that rely on ground-truth segmentation maps. Utilising semantic proportions suggested in this work offers a promising direction for future semantic segmentation research\footnote{Code available at \url{https://github.com/Halilibrahimaysel/Semantic_Segmentation_by_Semantic_Proportions}}.
\end{abstract}

%---------------------

%---------------------


%---------------------

\section{Introduction}
\IEEEPARstart{S}{emantic} segmentation is the task of partitioning an image into different regions depending on their semantic classes/categories. It
is widely used in a variety of fields such as autonomous driving \cite{siam2018comparative}, medical imaging \cite{asgari2021deep, yang2021artificial}, augmented reality \cite{zhang2020slimmer} and robotics \cite{milioto2018real}. Impressive
improvements have been shown in those areas with the recent development of deep neural networks (DNNs), benefiting from the availability of extensive annotated segmentation datasets at a large scale \cite{garcia2017review, hao2020brief}. However, creating such datasets can be expensive and time-consuming due to the usual need to annotate pixel-wise labels as it takes between 54 and 79 seconds per object \cite{bearman2016s}, thus requiring a couple of minutes per image with a few objects.  Moreover, requiring full supervision is rather impractical in some cases, for example, in medical imaging where expert knowledge is required. Annotating 3D data for semantic segmentation is even more costly and time-consuming due to the additional complexity and dimensionality of the data, which generally requires voxel (i.e., point in 3D space) annotation. Skilled annotators from outsourcing companies that are dedicated to data annotation may be needed for specific requests to ensure annotation accuracy and consistency, adding further to the cost \cite{genova2021learning}. In addition, saving the annotated data could also be expensive given the substantial amount of storage space generally needed. 

\begin{figure*}[!h]
    \centering
    \includegraphics[width=0.88\textwidth,height=0.25\textwidth]{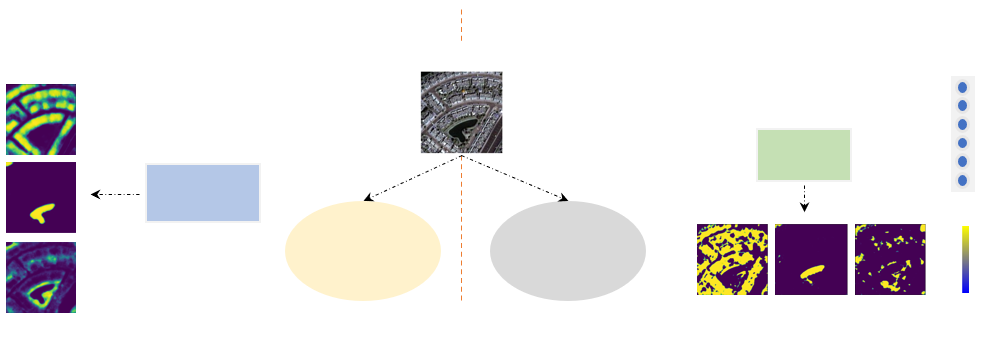}
    \put(-385, 111){ \textbf{Benchmark methods}}
    \put(-255, 105){\small {\scriptsize Input: $\vect{X}_i$}}
    \put(-309, 7){\scriptsize {Annotation: $\vect{Y}^*_{i}$}}
    \put(-311, 40){\scriptsize \textit{point supervision}}
    \put(-315, 33){\scriptsize \textit{segmentation mask}}
    \put(-290, 30){.}
    \put(-290, 28){.}
    \put(-290, 26){.}
    \put(-303, 19){\scriptsize \textit{ scribbles}}
    \put(-377, 58){\scriptsize Benchmark}
    \put(-372, 51){\scriptsize models}
    \put(-464, 101){\scriptsize \textit{Predicted masks: $\vect{Y}_{i}$}}
    \put(-325, 58){{$\xleftarrow{\hspace{1cm}}$}}
    \put(-180, 66){{$\xrightarrow{\hspace{1.5cm}}$}}
    \put(-55, 66){{$\xrightarrow{\hspace{0.7cm}}$}}
    \put(-140, 111){\textbf{Ours}}
    \put(-199, 7){\scriptsize {SP: $\vect{\rho}^*_{i}$}}
    \put(-34, 104){\scriptsize \textit{Predicted SP: {\scriptsize \textit{{$\vect{\rho}_{i}$}}}}} 
    \put(-110, 8){\scriptsize \textit{Predicted masks: $\vect{Y}_i$}}
    \put(-213, 42){\scriptsize \textit{building}}
    \put(-185, 42){\scriptsize \textit{: 68\%}}
    \put(-211, 35){\scriptsize \textit{water}}
    \put(-185, 35){\scriptsize \textit{: 4\%}}
    \put(-197, 32){.}
    \put(-197, 30){.}
    \put(-197, 28){.}
    \put(-216, 21){\scriptsize \textit{vegetation}}
    \put(-185, 21){\scriptsize \textit{: 7\%}}
    \put(-101, 69){\scriptsize Our model}
    \put(-8, 39){\scriptsize 1}
    \put(-8, 19){\scriptsize 0}
    \caption{Difference between the proposed semantic segmentation approach and benchmark methods.}
    \label{fig:first_fig}
\end{figure*}

Different approaches have been proposed to reduce the fine-grained level (e.g. pixel-wise) annotation costs. One line of research is to train segmentation models in a weakly supervised manner by requiring image-level labels \cite{pinheiro2015image, wei2016learning}, scribbles \cite{lin2016scribblesup}, eye tracks \cite{papadopoulos2014training}, or point supervision \cite{bearman2016s, mcever2020pcams} rather than costly segmentation masks of individual semantic classes. 
In contrast, in this paper we propose to utilise the proportion (i.e., percentage information) of each semantic class present in the image for semantic segmentation. For simplicity, we call this type of annotation {\it semantic (class) proportions} (SP). To the best of our knowledge, this is the first time of utilising SP for semantic segmentation. This innovative way, different from the existing ways (see e.g. Figure \ref{fig:first_fig}), could significantly simplify and reduce the human involvement required for data annotation and storage space in semantic segmentation. Our proposed approach by utilising the SP annotation can achieve comparable and sometimes even better performance in comparison to benchmark methods with full supervision utilising ground-truth segmentation masks. Moreover, we show that our method can sometimes provide free performance improvement in the presence of  ground-truth maps as it can be served as a plug-and-play module, which can easily be added on top of existing DNNs trained for segmentation tasks.

Our main contributions are: i) propose a new semantic segmentation methodology and a plug-and-play module, utilising SP annotations; ii) conduct extensive experiments on representative benchmark datasets from distinct fields to demonstrate the effectiveness and robustness of the proposed approach; and iii) draw an insightful discussion for semantic segmentation with weakly annotated data and future directions.

%---------------------
\section{Related Work}
\label{related_work}
%---------------------
\textit{Supervision levels in semantic segmentation}. In recent years, more and more researchers have focused on reducing the annotation cost for semantic segmentation tasks. One way is to use weakly supervised learning techniques that require less precise or less expensive forms of supervision. For instance, the work in \cite{wei2016learning} proposed to utilise image-level labels, the work in \cite{papandreou2015weakly,dai2015boxsup} used bounding boxes, and the methods in \cite{lin2016scribblesup, lee2020scribble2label} fed scribbles as labels instead of precise annotations to conduct semantic segmentation. Those approaches can significantly reduce the annotation cost, as they require less manual effort to annotate the data. However, there is always a trade-off between the annotation cost and the model performance, i.e., models trained with higher levels of supervision generally perform better than weakly supervised models.
Active learning is an alternative approach to reduce the annotation cost by selecting the most informative samples to annotate based on the current model's uncertainty. With the selected most informative samples, active learning can reduce the amount of data that needs to be labelled, thus reducing the annotation cost \cite{mittal2023revisiting, xie2020deal}. It is worth mentioning that this is actually similar to the way we propose for the SP degraded by clustering presented in Section \ref{subsec:SP_cluster}. Reducing the annotation cost could also be achieved by generating synthetic data that can be used to augment the real-world data \cite{chen2019learning}. Synthetic data can be generated using e.g. computer graphics or other techniques to simulate realistic images and labels.

\textit{DNNs for semantic segmentation}. The work in \cite{long2015fully}  made a breakthrough by proposing fully convolutional networks (FCNs) for semantic segmentation. FCNs utilise convolutional neural network (CNN) to transform input images into a probability map, where each entry of the probability map represents the likelihood of the corresponding image pixel belonging to a particular class. This approach allows the model to learn spatial features and eliminate the need for hand-crafted features. Following FCN, several variants have been proposed to improve the segmentation performance. For example, SegNet \cite{badrinarayanan2017segnet} is a modification of FCN employing an encoder-decoder architecture to achieve better performance; and DeepLab \cite{chen2017deeplab} introduced a novel technique called atrous spatial pyramid pooling to capture multi-scale information from the input image. 
U-Net \cite{ronneberger2015u}, one of the architectures used in our proposed methodology, is a type of CNN consisting of a contracting path and an expansive path. 
The skip connections in U-Net allow the network to retain and reuse high-level feature representations learned in the contracting path, helping to improve segmentation accuracy.
The U-Net architecture has been widely used for biomedical image segmentation tasks such as cell segmentation \cite{hu2019mc}, organ segmentation \cite{chen2023multi} and lesion detection \cite{dildar2021skin, cao2021automatic}, due to its ability to accurately segment objects within images while using relatively few training samples. Furthermore, its modular architecture and efficient training make it adaptable to a wide range of segmentation tasks. Therefore, to demonstrate our methodology utilising SP, we employ a modified and relatively basic version of the U-Net architecture as the backbone of our models for most of the experiments.

%---------------------
\section{Methodology}
\label{sec:methodology}
%---------------------
{\it Notation}. Let ${\cal X}$ be a set of images. Without loss of generality, we assume each image in ${\cal X}$ contains no more than $C$ semantic classes. $\forall \vect{X}_i \in {\cal X}$, $\vect{X}_i \in \mathbb{R}^{M\times H}$, where $M\times H$ is the image size. Let ${\cal X}_{\rm T} \subset {\cal X}$ and ${\cal X}_{\rm V} \subset {\cal X}$ be the training and validation (test) sets, respectively; and let ${\Omega}_{\rm T} \subset \mathbb{N}$ be the set containing the indexes of the images in ${\cal X}_{\rm T}$.
$\forall \vect{X}_i \in {\cal X}_{\rm T}$, annotations are available. The most general annotation is the ground-truth segmentation maps, say $\{\vect{Y}^*_{ij}\}_{j=1}^C$, for $\vect{X}_i$, where each $\vect{Y}^*_{ij} \in \mathbb{R}^{M\times H}$ is a binary mask for the semantic class $j$ of $\vect{X}_i$. For simplicity, let $\vect{Y}^*_{i}$ be a tensor formed by $\{\vect{Y}^*_{ij}\}_{j=1}^C$, where its $j$-th channel is $\vect{Y}^*_{ij}$.
Note that the ground-truth segmentation maps are not required in our approach for semantic segmentation in this paper unless specifically stated; instead, they are mainly used by benchmark methods for the comparison purpose. Analogously, let $\vect{Y}_{i}$ be the predicted segmentation maps following the same format as $\vect{Y}^*_{i}$.
Let $\vect{\rho}^*_{i} = (\rho^*_{i1}, \cdots, \rho^*_{iC})$ be the given SP annotation of image $\vect{X}_i \in {\cal X}_{\rm T}$, which will be mainly used to train our approach, where each $\rho^*_{ij} \in [0, 1]$ is the SP of the $j$-th semantic class of $\vect{X}_i$ and $\sum_{j=1}^C \rho^*_{ij} = 1$.

\begin{figure*}[!h]
    \centering    \includegraphics[width=0.8\textwidth,height=0.15\textwidth]{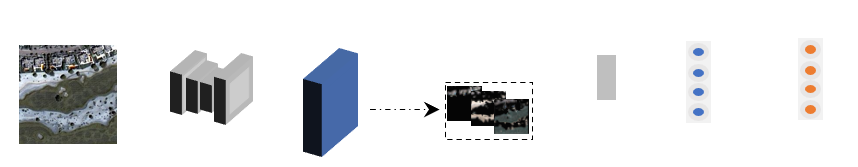}
     \put(-394, 61){\scriptsize {Input: $\vect{X}_i$}}
    \put(-341, 8){\scriptsize \textbf{Feature extraction}}
     \put(-343, 13){$\xrightarrow{\hspace{2cm}}$}
    \put(-314, 61){\scriptsize \textit{CNN}}
    \put(-264, 61){\scriptsize {Features: $\vect{Y}_i$}}
    \put(-214, 4){\scriptsize Predicted segmentation maps}
    \put(-142, 25){\scriptsize \textbf{SP computation}}
    \put(-225, 42){$\xrightarrow{\hspace{3cm}}$}
    \put(-112, 38){$\xrightarrow{\hspace{0.5cm}}$}
    \put(-228, 18){\scriptsize i.e.,}
    \put(-127, 55){\scriptsize \textit{GAP}}
    \put(-57, 45){\small $\mathcal{L}_{\rm sp}$}
    \put(-59, 35){$\longleftrightarrow$}
    \put(-95, 12){\scriptsize Predictions: $\vect{\rho}_{i}$}
    \put(-42, 12){\scriptsize Ground-truth: $\vect{\rho}^*_{i}$}
    \caption{The SPSS (SP-based semantic segmentation) architecture. In the training stage, features are firstly extracted by a CNN from the input; and then the extracted features are through a GAP layer calculating the SP. After training using the loss function $\mathcal{L}_{\rm sp}$, the proposed SPSS architecture can force the extracted features to be the prediction of the class-wise segmentation masks. }
    \label{fig:model1}
\end{figure*}

%%%%%%%
\begin{figure*}[!h]
    \centering  \includegraphics[width=0.9\textwidth,height=0.22\textwidth]{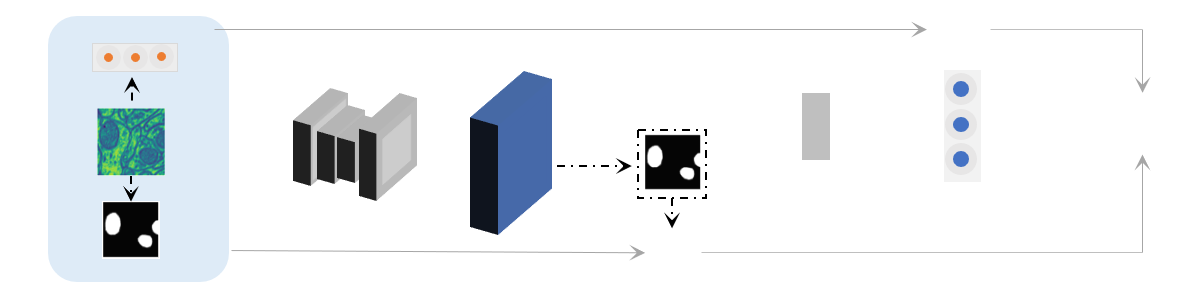}
    \put(-365, 95){\scriptsize \textit{Loss for SP}}
    \put(-433, 100){\scriptsize {Annotation: \textit{\blue{SP}}}}
    \put(-433, 10){\scriptsize Annotation: \textit{\blue{SM}}}
     \put(-357, 32){\scriptsize \textbf{Feature extraction}}
     \put(-360, 25){$\xrightarrow{\hspace{2cm}}$}
     \put(-270, 90){\scriptsize Features}
    \put(-170, 46){\scriptsize \textbf{SP computation}}
    \put(-240, 70){$\xrightarrow{\hspace{2.5cm}}$}
    \put(-140, 65){$\xrightarrow{\hspace{1cm}}$}
    \put(-365, 10)
    {\scriptsize \textit{Loss for semantic SM (segmentation map)}}
    \put(-327, 80){\scriptsize \textit{CNN}}
    \put(-153, 82){\scriptsize \textit{GAP}}
    \put(-93, 92){$\scalebox{1}{$\uparrow$}$}
    \put(-97, 103){\small $\mathcal{L}_{\rm sp}$}
    \put(-209, 15){\small $\mathcal{L}_{\rm sm}$}
    \put(-32, 71){\small $\mathcal{L}_{\rm total}=$}
    \put(-44, 59){\small $\alpha \mathcal{L}_{\rm sp}+(1-\alpha) \mathcal{L}_{\rm sm}$}
    \caption{The SPSS+ architecture ({\it cf.} the SPSS architecture in Figure \ref{fig:model1}). In contrast, $\mathcal{L}_{\rm total}$ (see Eq. \eqref{eqn:loss-total}), a weighted average of $\mathcal{L}_{\rm sp}$ and $\mathcal{L}_{\rm sm}$,  is calculated during training. After training, the SPSS+ architecture can force the extracted features to be the prediction of the class-wise segmentation masks.} 
    
    \label{fig:model2}
\end{figure*}

{\it Loss function}. Two types of loss functions are introduced in the architectures of our method. One is based on the mean squared error (MSE). MSE is commonly used to evaluate the performance of regression models where there are numerical target values to predict. We employ MSE to measure the discrepancy between the ground-truth SP and the predicted ones. For ease of reference, we call this loss function $\mathcal{L}_{\rm sp}$ throughout the paper, i.e.,
\begin{equation} \label{eqn:loss-sp}
\mathcal{L}_{\rm sp} = \frac{1}{|{\Omega}_{\rm T}|} \sum_{i\in {\Omega}_{\rm T}}\|\vect{\rho}^*_{i} - \vect{\rho}_{i}\|^2,
\end{equation}
where $\vect{\rho}_{i}$ is the predicted SP for image $\vect{X}_i \in {\cal X}_{\rm T}$ and $|{\Omega}_{\rm T}|$ is the cardinality of set ${\Omega}_{\rm T}$. 
The other loss function, which will be deferred in Section \ref{subsec:spss+}, is defined based on the binary cross-entropy (BCE). BCE is a commonly used loss function in binary classification problems and measures the discrepancy between the predicted probabilities and the true binary ones. Below we define the BCE function as 
\begin{equation} \label{eqn:loss-bce}
\begin{split}
\mathcal{L}_{\rm sm}  = & \frac{1}{|{\Omega}_{\rm T}|} \sum_{i\in {\Omega}_{\rm T}} \sum_{j=1}^C - (\vect{Y}^*_{ij} \log(\vect{Y}_{ij}) \ + \\
& \ (1- \vect{Y}^*_{ij}) \log(1 - \vect{Y}_{ij})),
\end{split}
\end{equation}
where $\vect{Y}_{ij}$ is the predicted segmentation map for the $j$-th semantic class of image $\vect{X}_i\in {\cal X}_{\rm T}$.

\subsection{Proposed SP-based Semantic Segmentation Architecture} 
%-------
The proposed SP-based semantic segmentation (SPSS) architecture is shown in Figure \ref{fig:model1}. It contains two main parts.
The first part of the SPSS architecture is feature extraction. 
Employing a CNN is a common approach in current state-of-the-art semantic segmentation methods. In our SPSS, a CNN (or other type of DNNs) is utilised as its backbone to extract high-level image features $\vect{Y}_i$ from the input image $\vect{X}_i$. The second part of the SPSS architecture is a global average pooling (GAP) layer, which takes the image features $\vect{Y}_i$ to generate the SP, $\vect{\rho}_{i}$, for the input image $\vect{X}_i$. The SPSS architecture is then trained by using the loss function $\mathcal{L}_{\rm sp}$ defined in Eq. \eqref{eqn:loss-sp}. After training the SPSS architecture, the extracted features $\vect{Y}_i$ of the trained CNN are, surprisingly, the prediction of the class-wise segmentation masks; that is how the SPSS architecture performs semantic segmentation by just using the SP rather than the ground-truth segmentation maps. 

We remark that both parts in the SPSS architecture except for utilising SP are well-known and commonly employed for e.g. computer vision tasks. To the best of our knowledge, it is, for the first time, to combine them for semantic segmentation in reducing the need of labour-intensive (fine-grained) ground-truth segmentation masks to the (coarse-grained) SP level.

%-------
\subsection{A Booster: SPSS+} 
\label{subsec:spss+}

The proposed SPSS architecture in Figure \ref{fig:model1} only uses the SP annotation for semantic segmentation, which is quite cheap in terms of annotation generation. Moreover, SPSS is also very flexible. For example, i) the proposed loss function $\mathcal{L}_{\rm sp}$ using SP can be employed as a plug-and-play module in different DNNs; and ii) SPSS can be enhanced directly when additional annotation information is available. Below we give a showcase regarding how to use SP and pixel-level annotations jointly to enhance the SPSS architecture, see Figure \ref{fig:model2}. For ease of reference, we call the proposed booster in Figure \ref{fig:model2} {\it SPSS+}.

The total loss for the SPSS+ architecture is 
\begin{equation} \label{eqn:loss-total}
\mathcal{L}_{\rm total} = \alpha \mathcal{L}_{\rm sp} +(1-\alpha) \mathcal{L}_{\rm sm},
\end{equation}
where $\alpha$ is an adjustable weight to determine the trade-off between  $\mathcal{L}_{\rm sp}$ and  $\mathcal{L}_{\rm sm}$.
The SPSS+ architecture uses the loss $\mathcal{L}_{\rm total}$, which considers the annotations of the SP and segmentation masks for training. Similar to the SPSS architecture (in Figure \ref{fig:model1}),
the extracted features $\vect{Y}_i$ of the trained CNN in the SPSS+ architecture are the prediction of the class-wise segmentation masks, i.e., the semantic segmentation results.

Our SPSS can generally achieve comparable performance against benchmark semantic segmentation methods. SPSS+ works as a performance booster and improves the segmentation ability of SPSS without extra training data or model complexity. More details regarding the extensive validation and comparison are given in Section \ref{sec:experiments}.

\begin{figure*}[!h
tb]
    \centering \includegraphics[width=0.85\textwidth,height=0.17\textwidth]{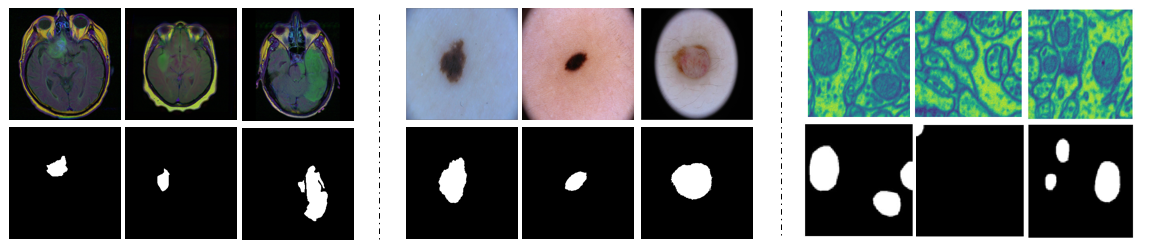}
    \put(-410, 90){\tt LGG Brain MRI}
    \put(-231, 90){\tt ISIC}
    \put(-126, 90){\tt Electron Microscopy}
    \caption{Example images and ground-truth segmentation masks of the three employed medical imaging datasets.}
    \label{fig:medical_datasets}
\end{figure*}

%---------------------
\section{Data and Settings} \label{sec:data-setting}
%---------------------

\subsection{Data}

The proposed SP-based methodology for semantic segmentation is showcased on four different datasets described below.

\textit{(i)} Satellite images of Dubai, i.e., {\tt Aerial Dubai}. This is an open-source aerial imagery dataset presented as part of a Kaggle competition\footnote{{\url{https://www.kaggle.com/datasets/humansintheloop/semantic-segmentation-of-aerial-imagery}}}. The dataset includes 8 tiles and each tile has 9 images of various sizes and their corresponding ground-truth segmentation masks for 6 classes, \textit{i.e., building, land, road, vegetation, water and unlabeled}.

\textit{(ii)} Medical imaging dataset {\tt ISIC} (International Skin Imaging Collaboration). This is a comprehensive collection of dermoscopic images specifically curated for the study and analysis of skin lesions \cite{codella2019skin, tschandl2018ham10000}. It contains 2594 training, 100 validation and 1,000 test  images with high-resolution capturing various types of skin lesions, including benign and malignant conditions. Each image in the dataset is accompanied by expert annotations including detailed segmentation masks outlining the precise boundaries of the lesions. These annotations are crucial for segmentation methods to accurately delineate the lesion from the surrounding skin. The {\tt ISIC} dataset is frequently used in research and competitions, such as the {\tt ISIC} Challenge, to benchmark and advance segmentation algorithms. However, obtaining fine-grained pixel-level segmentation masks is expensive and our SPSS model shows comparable performance despite being trained with dramatically less expensive SP rather than full masks in Section IV in the main paper.

\textit{(iii)} Medical imaging dataset {\tt Electron Microscopy}\footnote{{\url{https://www.epfl.ch/labs/cvlab/data/data-em/}}}. It contains 165 slices of microscopy images with the size of $768\times1024$. The primary aim of this medical dataset is to identify and classify mitochondria pixels. 
This dataset is quite challenging since its semantic classes are severely imbalanced, i.e., the size of the mitochondria in most slices is very small (e.g. see the right column of Figure \ref{fig:medical_datasets} and Figure \ref{fig:mic_better}). 

\textit{(iv)} Medical imaging dataset {\tt LGG Brain MRI} from The Cancer Genome Atlas (TCGA) and The Cancer Imaging Archive (TCIA). We used the version made available by Buda et al. \cite{buda2019association} on Kaggle\footnote{{\url{https://www.kaggle.com/datasets/mateuszbuda/lgg-mri-segmentation}}}, where the authors selected 120 patients from the TCGA lower-grade glioma collection\footnote{{\url{https://cancergenome.nih.gov/cancersselected/lowergradeglioma}}} which had available preoperative imaging data including at least a fluid-attenuated inversion recovery (FLAIR) sequence. The dataset includes roughly 4000 brain MRI images of 110 patients from 5 institutions.
Figure \ref{fig:medical_datasets} presents some example images for the three medical imaging datasets.

\subsubsection{Data Preprocessing}

The {\tt Aerial Dubai} and {\tt Electron Microscopy} datasets contain large images that were preprocessed into smaller patches for analysis. Specifically, each image in the {\tt Aerial Dubai} dataset was divided into $224\times224$ pixel patches, resulting in a total of 1,647 images. For the {\tt Electron Microscopy} dataset, images were divided into $256 \times 256$ pixel patches, yielding 1,980 images.  The images in the {\tt LGG Brain MRI} dataset, originally sized at $256\times256$ pixels, were centre-cropped to $144\times144$ pixels. Subsequently, images from all datasets including {\tt ISIC} were then resized to $288 \times 288$ pixels. This preprocessing ensures uniformity in image sizes across different datasets, facilitating consistent and effective analysis.

\subsection{Experimental Settings}

Benchmark methods with different CNN backbones (e.g., U-Net \cite{ronneberger2015u} or Feature Pyramid Network (FPN) \cite{lin2017feature} with VGG16 \cite{simonyan2014very} and ResNet34 \cite{he2016deep}) are trained end-to-end for semantic segmentation using the ground-truth segmentation masks, comparing to ours using the SP. For fair comparison, the same training images are used to train all the models.

\subsubsection{Deep Neural Architecture Details}

\begin{itemize}
\item We employed U-Net \cite{ronneberger2015u} and FPN \cite{lin2017feature} architectures with pre-trained weights from VGG16 \cite{simonyan2014very} and ResNet34 \cite{he2016deep} on the {\tt Aerial Dubai} dataset. For the medical imaging datasets and all the ablation experiments presented in Section \ref{sec:experiments}, we consistently utilized a U-Net with VGG16 weights.

\item To adapt U-Net and FPN for predicting SP rather than fine-grained masks, a $1\times1$ convolutional layer with $n$ filters is employed to match the $C$ number of the semantic classes. 
Thus $n$ is set to 6 and 1 to output feature maps of the size $288\times288\times6$ and $288\times288\times1$ respectively for the {\tt Aerial Dubai} and medical imaging datasets. Note that there is no need to set $n$ to 2 for the binary segmentation problem with medical imaging datasets. Finally, a global average pooling (GAP) layer added on top to get $n$ float to be used as the predicted SP values.

\item To obtain segmentation maps during the test stage, we extract the feature maps prior to the GAP layer and visualise them per semantic class ({\it cf.} Figures \ref{fig:model1} and \ref{fig:model2}). 

\end{itemize}

\subsubsection{Training Setup}

For all experiments, an 80/20 split for the training/test, Adam optimizer with a learning rate of $10^{-3}$, and a batch size of 16 were chosen. The number of epochs was set to 100 with early stopping applied with patience set to 10 based on the validation loss. 
All the experiments were implemented on a personal laptop with the following specifications: i7-8750H CPU, GeForce GTX 1060 GPU and 16GB RAM.
Training of SPSS and SPSS+ takes around  30 minutes and 40 minutes, respectively.

\begin{figure*}[!htb]
    \centering    \includegraphics[width=0.9\textwidth,height=0.16\textwidth]{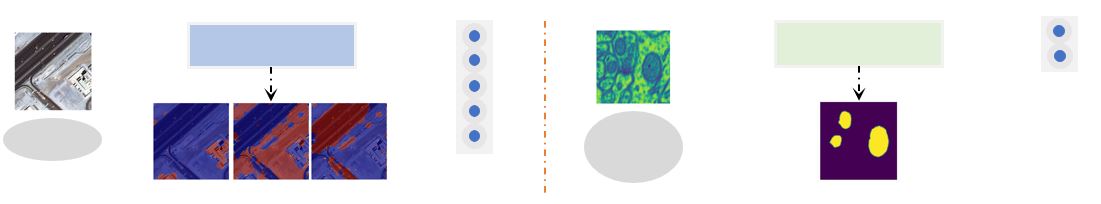}
    \put(-455,72){\scriptsize {Input: $\vect{X}_i$}}
    \put(-459, 10){\scriptsize Annotation}
    \put(-452, 24){\scriptsize SP: $\vect{\rho}^*_{i}$}
    \put(-373, 61){\small {SPSS Model}}
    \put(-212,73){\scriptsize {Input: $\vect{X}_i$}}
    \put(-212, 1){\scriptsize Annotation}
    \put(-206, 28){\scriptsize SP: $\vect{\rho}^*_{i}$}
    \put(-200, 20){\scriptsize \&}
    \put(-210, 12){\scriptsize SM: $\vect{Y}^*_{i}$}
    \put(-405, 60){$\xrightarrow{\hspace{0.5cm}}$}
    \put(-295, 60){$\xrightarrow{\hspace{0.5cm}}$}
    \put(-274, 12){\scriptsize SP:  $\vect{\rho}_{i}$}
    \put(-165, 60){$\xrightarrow{\hspace{0.5cm}}$}
    \put(-55, 60){$\xrightarrow{\hspace{0.5cm}}$}
    \put(-27, 44){\scriptsize SP: $\vect{\rho}_{i}$}
    \put(-345, 46){\scriptsize Segmentation map}
    \put(-397, 3){\scriptsize building}
    \put(-355, 3){\scriptsize land}
    \put(-325, 3){\scriptsize road}
    \put(-128, 62){\small {SPSS+ Model}}
    \put(-95, 46){\scriptsize Segmentation map}
    \put(-121, 2){\scriptsize mitochondria}
    \caption{Diagrams of the proposed models SPSS and SPSS+ on the datasets {\tt Aerial Dubai} (\textit{left}) and {\tt Electronic Microscopy} (\textit{right}; significant class imbalance), respectively.
    }
    \label{fig:difference}
\end{figure*}

%---------------------
\section{Experiments} \label{sec:experiments}
%---------------------

\begin{table*}[!ht]
\centering
\caption{Quantitative semantic segmentation results (Mean IoU and F1 scores) on the \texttt{Aerial Dubai} dataset.}
{\small 
\begin{tabular}{c|cc|cc|cc|cc} \hline 
{Model} & \multicolumn{4}{c|}{U-Net} & \multicolumn{4}{c}{FPN} \\
\cline{2-8} 
{Backbone} & \multicolumn{2}{c|}{VGG16} & \multicolumn{2}{c|}{ResNet34 } & \multicolumn{2}{c|}{VGG16} & \multicolumn{2}{c}{ResNet34}  \\
\hline
{Metric} & {Mean IoU} & {F1}  & {Mean IoU} & {F1} & {Mean IoU} & {F1}  & {Mean IoU} & {F1}  \\
\hline \hline
{Benchmark} & $71.3 \pm 1.2$ & $88.3 \pm 0.7$ & $69.2 \pm 0.8$ & $86.1 \pm 1.2$ & $\textbf{68.5} \pm \textbf{0.5}$ & $\textbf{82.1} \pm \textbf{0.3}$ & $67.2 \pm 0.8$  & $81.3 \pm 0.8$ \\
{SPSS} & $64.2 \pm 0.6$ & $83.7 \pm 0.4$ & $64.4 \pm 0.4 $ & $80.6 \pm 0.8$ & $60.5 \pm 0.2$ & $77.2 \pm 0.4$ & $61.7 \pm 0.6$ & $77.5 \pm 1.1$\\
{SPSS+} & $\textbf{71.6} \pm \textbf{0.6}$ & $\textbf{88.7} \pm \textbf{0.6}$ & $\textbf{70.4} \pm \textbf{0.5}$ & $\textbf{86.4} \pm \textbf{0.3}$ & $67.7 \pm 1.2$ & $80.5 \pm 0.5$ & $\textbf{69.2} \pm \textbf{1.0}$ & $\textbf{82.5} \pm \textbf{0.7}$\\
\hline
\end{tabular}
}
\label{table:Aerial_table}
\end{table*}

We highlight that the main aim here is to show that semantic segmentation can be achieved with significantly weaker annotations, i.e., the SP annotation, rather than segmentation accuracy enhancement only.
Recall that the difference between SPSS and SPSS+ is just the way of using the annotations for their training, i.e., SPSS+ addresses scenarios that ground-truth segmentation maps are available. Figure \ref{fig:difference} illustrates the difference by utilising the SPSS and SPSS+ architectures on the datasets {\tt Aerial Dubai} and {\tt Electronic Microscopy}, respectively.
To demonstrate the effectiveness of our semantic segmentation approach, we evaluate performance using mean Intersection over Union (IoU) and F1 scores.

%------
\subsection{Segmentation Performance Comparison}
\label{sec:results}
%------
\textit{Quantitative comparison}. Tables \ref{table:Aerial_table} and  \ref{table:results_medical} give the quantitative results of our method and the benchmark methods for the {\tt Aerial Dubai} and the three medical imaging datasets, respectively. Well-known evaluation metrics, i.e., mean intersection over union (Mean IoU) and F1 scores are employed. Estimated errors in the mean are obtained by training the models three times with randomly initialised weights.
Tables \ref{table:Aerial_table} and \ref{table:results_medical} show that SPSS performs comparably to the benchmark methods for all tasks, demonstrating the utility of the SP annotation for semantic segmentation that our methodology introduces. 
Moreover, SPSS+, i.e., using both ground-truth maps and SP, outperforms the benchmark methods for all the cases except for using the FPN with VGG16 backbone, indicating the usefulness of involving the SP annotation. Note again that SPSS+ does not require any additional data collection or increase in model complexity, hence  offering performance improvements for semantic segmentation tasks nearly for free.
Without loss of generality, U-Net with VGG16 is adopted in our method for the rest of the experiments.

\begin{figure*}[!h
t]
    \centering \includegraphics[width=0.8\textwidth,height=0.3\textwidth]{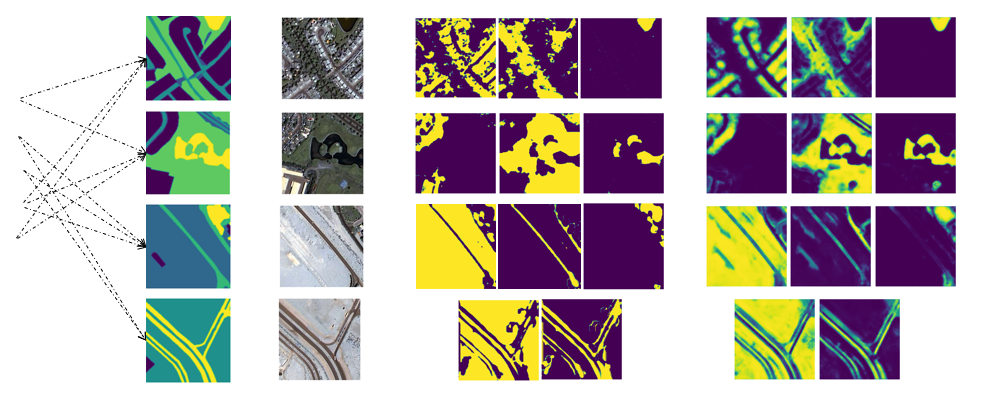}
    \put(-435, 127){\small \textbf{Classes:}}
    \put(-435, 115){\small building}
    \put(-425, 101){\small land}
    \put(-425, 87){\small road}
    \put(-445, 74) {\small vegetation}
    \put(-430, 62) {\small water}
    \put(-435, 30){\small \textbf{Dataset:}}
    \put(-435, 18){\small \texttt{Aerial Dubai}}
    \put(-357, 152){\small Ground truth}
    \put(-296, 152){\small {Input: $\vect{X}_i$}}
    \put(-210, 152){\small \textbf{SPSS} (ours)}
    \put(-105, 152){\small Benchmark method}
    \caption{Qualitative semantic segmentation comparison between our SPSS method (\textit{middle}) and the benchmark method (\textit{right}).}
    \label{fig:further}
\end{figure*}

\begin{table}[!ht]
\centering
\caption{Quantitative semantic segmentation results (Mean IoU scores) on the medical imaging datasets using U-Net with VGG16 backbone.}
\label{table:medical}
\resizebox{0.48\textwidth}{!}
{\small 
\begin{tabular}{c||c|c|c} \hline
\backslashbox{Method} {Data}  & {\tt ISIC} & {\tt  Mithocondria} & {\tt  Brain MRI} \\ \hline \hline
{Benchmark} & {$78.4 \pm 0.3$}  & {$83.7 \pm 0.6$}  &{$72.3 \pm 0.2$}  \\ 
SPSS        & {$73.2 \pm 0.5$}  & {$76.5 \pm 0.2$}  &{$69.5 \pm 0.6$}  \\ 
SPSS+       & {$\textbf{79.1} \pm \textbf{0.1}$}  & {$\textbf{84.3} \pm \textbf{0.5}$}  &{$\textbf{72.8} \pm \textbf{0.4}$}  \\ \hline
 \end{tabular}
}
\label{table:results_medical}
\end{table}

\begin{figure}[!h]
    \centering
    \includegraphics[trim={{.10\linewidth} {.00\linewidth} {.00\linewidth} {.00\linewidth}}, clip, width=0.51\textwidth, height=0.17\textwidth]{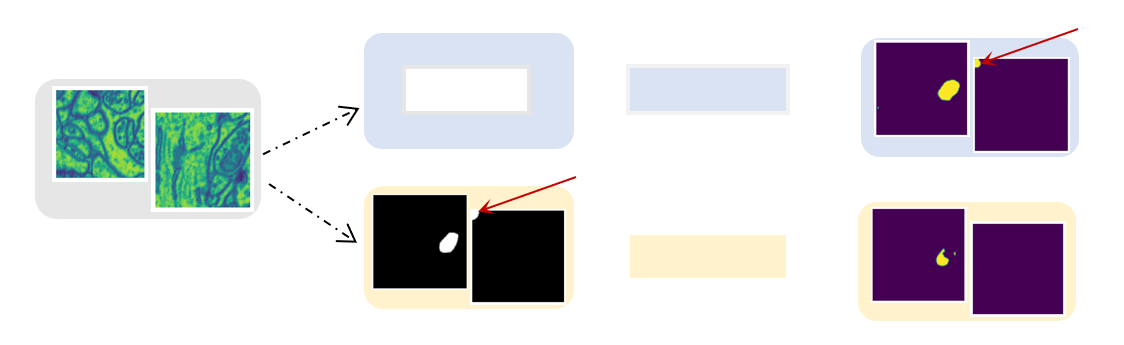}
    \put(-247, 70){\scriptsize {Images}}
    \put(-177, 82){\scriptsize {Annotations}}
    \put(-58, 83){\scriptsize {Predicted maps}}
    \put(-170, 63){\scriptsize  SP: $\vect{\rho}^*_{i}$}
    \put(-135, 63){$\xrightarrow{\hspace{0.3cm}}$}
    \put(-135, 20){$\xrightarrow{\hspace{0.3cm}}$}
    \put(-117,63){\scriptsize \textbf{Our SPSS+}}
     \put(-117,21){\scriptsize {Benchmark}}
    \put(-80, 63){$\xrightarrow{\hspace{0.3cm}}$}
    \put(-80, 20){$\xrightarrow{\hspace{0.3cm}}$}
     \caption{Comparison between our SPSS+ method (\textit{upper}) and the benchmark method (\textit{lower}) on some images from the \texttt{Electronic Microscopy} dataset. }
    \label{fig:mic_better}
\end{figure}

\textit{Qualitative comparison}. Figure \ref{fig:further} shows the qualitative results of our method and the benchmark method for the {\tt Aerial Dubai} dataset. Surprisingly, the class-wise segmentation maps that our method achieves (middle of Figure \ref{fig:further}) are visually significantly better than that of the benchmark method (right of Figure \ref{fig:further}) in terms of the binarisation ability, indicating the effectiveness of the loss $\mathcal{L}_{\rm sp}$ (defined in Eq. \eqref{eqn:loss-sp}) using the SP annotation we introduce. 
For the significant class imbalance dataset {\tt Electronic Microscopy}, Figure \ref{fig:mic_better} shows the qualitative results of our method and the benchmark method for some challenging cases. Again, our method exhibits superior performance against the benchmark method. For example, our method can accurately segment the mitochondria on the top-left corner of the second image despite employing much less annotation, but the benchmark method completely misses it despite being trained using the ground-truth segmentation masks. This again validates the effectiveness of the SP annotation for semantic segmentation.
Moreover, due to the great binarisation ability of the loss $\mathcal{L}_{\rm sp}$ using SP, it may serve as an auxiliary loss functioning as a plug-and-play module even in scenarios where ground-truth segmentation masks are available to enhance the segmentation performance of many existing methods as SPSS+ does.

\begin{figure*}[!h]
    \centering
    \includegraphics[width=0.85\textwidth, height=0.1\textwidth]{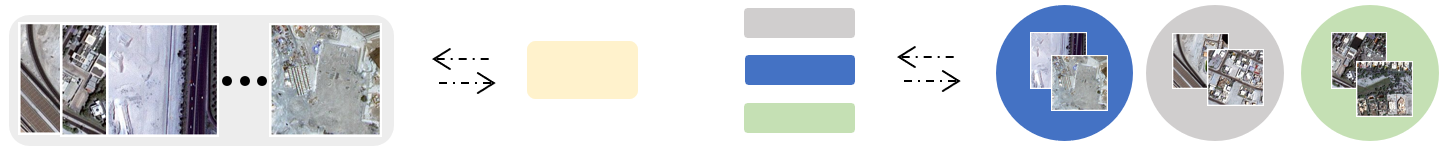}
    \put(-400, -7){\small {Training set}}
    \put(-275, -7){\small {SP} set}
    \put(-268, 25){\small \textbf{$\vect{\rho}^*_{i}$}}
    \put(-217, -7){\small {SP clusters}}
    \put(-235, 25) {\large \textcolor{blue}{$\longrightarrow$}}
    \put(-100, -7){\small {Image clusters}}
    \caption{Diagram of the SP annotation degraded by clustering. Images are clustered corresponding to the SP clusters which are achieved by applying $K$-means on the SP set. An SP annotation for one image in each image cluster is then randomly selected from that cluster and is assigned to all the images in that image cluster. 
    }
    \label{fig:sp-clustering}
\end{figure*}

%------------------
\subsection{Sensitivity Analysis}
\label{section:sensitivity}
%------------------
Obtaining precise SP annotations may be challenging and, as a result, annotators may provide rough estimates instead. We showcase that rough estimated SP is quite sufficient for our model to achieve good performance (further results are deferred in Section \ref{section:further-ana}). Below we first investigate the robustness of our models corresponding to the quality of the SP. Two extreme ways degrading the SP are examined: one is adding noises to the SP directly and the other is assigning images in individual clusters the same SP.

\subsubsection{SP degraded by different noise} \label{subsec:sp-de-noise}
We firstly conduct sensitivity analysis of our method SPSS by systematically adding Gaussian noise to the SP for the {\tt Aerial Dubai} dataset. Let $\mathcal{N}(0, \sigma)$ be the normal distribution with 0 mean and standard deviation $\sigma$. For the given SP $\vect{\rho}^*_{i} = (\rho^*_{i1}, \cdots, \rho^*_{iC})$ of $\forall \vect{X}_i \in {\cal X}_{\rm T}$, let $\vect{\tilde{\rho}}^*_{i} = (\tilde{\rho}^*_{i1}, \cdots, \tilde{\rho}^*_{iC})$, where
\begin{equation}
\tilde{\rho}^*_{ij} = \rho^*_{ij} + \mathcal{N}(0, \sigma), \ \ j = 1, \cdots, C.
\end{equation}
The above steps are also summarized in Algorithm \ref{Alg:noisy-SP} in Appendix.
Then the softmax operator is used to normalise $\vect{\tilde{\rho}}^*_{i}$, and the normalised $\vect{\tilde{\rho}}^*_{i}$ is used as the new SP to train our model. Here the standard deviation $\sigma$ controls the level of the Gaussian noise being added to the SP; e.g., $\sigma = 0.1$ represents $10\%$ Gaussian noise.
Table \ref{table:noisy_results} showcases the robustness of our methodology, as it continues performing well even with the SP degraded by quite high levels of noise. E.g., the Mean IoU our method suffers a drops in performance of $\sim 4\%$ for $10\%$ Gaussian noise being added to the SP. Our method still works significantly above random guessing even with the SP which is degraded by $50\%$ Gaussian noise. This shows that our method is quite robust corresponding to the SP, which means the annotators could in practice spend much less effort for providing rough SP rather than the precise SP.

For medical imaging datasets, the SP of the positive class region, i.e., $\rho^*_{i1}$, is degraded by a different noise generation process to present diverse noise injection scenarios. Noise is added in a controlled manner utilising the uniform distribution $\mathcal{U}(a, b)$ bounded by $a$ and $b$, ensuring that the degraded SP remains within a specified range, i.e.,
\begin{equation}
\tilde{\rho}^*_{i1} = \rho^*_{i1} + \lambda \ \mathcal{U}(a, b) \rho^*_{i1},
\end{equation}
where $\lambda$ is a parameter with value $-1$ or 1 selected randomly.
The above way ensures that the degraded SP is relative to the size of the original SP controlled by bounds $a$ and $b$. The above steps are also summarized in Algorithm \ref{Alg:noisy-SP} in Appendix.
The results presented in Table \ref{table:noisy_medical} again show that our method SPSS is robust against high level of noise imposed on the SP.

\begin{table}[!htb]
\vspace{0.0in}
\centering
\caption{{Performance of our model in terms of Mean IoU trained by using the SP degraded by Gaussian noise.} }
\resizebox{0.48\textwidth}{!}
{\small 
\begin{tabular}{c|cccccccc}
\hline 
      & \multicolumn{8}{c}{Dataset \ {\tt Aerial Dubai}}  \\ \hline  \hline 
{Noise (\%)} & 0 & 5 & 10 & 15 & 20 & 30 & 40 & 50 \\  
{Mean IoU}   & 64.2 & 62.4 & 60.1 & 57.8 & 52.2 & 48.3 & 43.4 & 38.3  \\ \hline
\end{tabular}
}
\label{table:noisy_results}
\end{table}

\begin{table}[!htb]
\centering
\caption{Performance of our model in terms of Mean IoU trained by using the degraded SP for medical imaging datasets.}
\resizebox{0.48\textwidth}{!}
{\small 
\begin{tabular}{c||c|c|c} \hline
\backslashbox{Noise ($[a, b]$)} {Data}  & {\tt ISIC} & {\tt Mithocondria} & {\tt Brain MRI} \\ \hline \hline
{Noise free} & {$73.2$}  & {$76.5$}  &{$69.5$}  \\ 
{$[0, 0.5]$}        & {$70.1$}  & {$70.5$}  &{$62.5$}  \\ 
{$[0, 1]$}      & {$67.3$}  & {$66.2$}  &{$60.1$}  \\ 
{$[0.5, 1]$}      & {$69.3$}  & {$64.2$}  &{$63.1$}  \\ \hline
 \end{tabular}
}
\label{table:noisy_medical}
\end{table}

\begin{table}[!htb]
\vspace{0.0in}
\centering
\caption{{Performance of our model in terms of Mean IoU trained by using the SP degraded by clustering.} }
{\small 
\begin{tabular}{c|cccccc}
\hline
 & \multicolumn{6}{c}{Dataset \ {\tt Aerial Dubai}} \\ \hline \hline
{\# Clusters} $K$ & 100 & 50 & 30 & 20 & 10 & 5 \\ 
{Mean IoU}    & 61.7  &  59.4 & 56.5 & 51.2 & 47.4 & 38.3 \\ \hline
\end{tabular}
}
\label{table:cluster}
\end{table}

%----
\begin{table}[!htb]
\centering
\caption{{Comparison between the annotation styles of obtaining the segmentation masks and the SP in terms of time and memory. The \texttt{Aerial Dubai} dataset is used.}}
\resizebox{.47\textwidth}{!}
{\small 
\begin{tabular}{c||c|cc}
\hline
{Annotation} &  {Average time} & \multicolumn{2}{c}{Memory per image}  \\   %\cline{3-8}
style & per image & {Original} & {Compressed}   \\
\hline \hline
\textit{Segmentation masks} & $\sim330$s   & $\sim148$ kB & $\sim4$ kB  \\ %\hline
 \textit{SP (via annotators)}     & $\sim20$s & \multicolumn{2}{c} {$\sim0.02$ kB} \\ 
\hline
\end{tabular}
}
\label{table:annotation_efficiency}
\end{table}

\subsubsection{SP degraded by clustering} \label{subsec:SP_cluster} We now conduct the sensitivity analysis of our method by degrading the SP of the training images by clustering. The degradation procedures are: i) clustering the set of the given SP, i.e., $\{\vect{\rho}^*_{i}\}_{i \in {\Omega}_{\rm T}}$, into $K$ clusters by $K$-means; ii) clustering the training set ${\cal X}_{\rm T}$ into the same $K$ clusters, say ${\cal X}_{\rm T}^k, k=1, \ldots, K$, corresponding to the SP clusters; and iii) assigning all the training images in cluster ${\cal X}_{\rm T}^k$ the same SP which is randomly selected from the SP of one image in this cluster; see also Figure \ref{fig:sp-clustering} for illustration. 
Obviously, implementing this way of degrading the SP, all the images' SP in the training set ${\cal X}_{\rm T}$ are changed except for $K$ (i.e., the number of clusters) images if every training image has different SP annotation in the original SP set. The smaller the number $K$, the severer the SP degradation.

The performance of our method regarding the SP degraded by clustering is shown in 
Table \ref{table:cluster}, indicating again the robustness of our methodology corresponding to the SP. For example, after just using $K=100$ images' SP for the whole training set ${\cal X}_{\rm T}$, the Mean IoU of our method only drops by $\sim 2.5\%$; and just using $K=5$ images' SP for the whole training set, our method can still work to some extent (i.e., the Mean IoU just drops less than half).  This again shows that our method is indeed quite robust corresponding to the SP. This suggests one possible strategy to reduce effort is to cluster images (for example from patients with a similar level of disease) and then estimate SP on represetive images in the cluster.

\begin{table*}[!h]
\centering
\caption{{Quantitative comparison on the \texttt{Aerial Dubai} dataset with rough estimated SP annotations.}}
{\small 
\begin{tabular}{c||c|ccccc|c}
\hline
\multirow{2}{*}{Model}  & \multirow{2}{*}{Mean IoU} & \multicolumn{5}{c|}{Per-class F1 score} & \multirow{2}{*}{Mean accuracy}  \\   %\cline{3-8}
  &  & {Building} & {Land} & Road & Vegetation & Water  &   \\
\hline \hline
 \textit{Segmentation masks} & $39.5 \pm 1.3$  & $\textbf{52.7} \pm \textbf{1.2}$ & $84.8 \pm 0.6$ & $2.4 \pm 0.6$ & $43.2 \pm 1.3$ & $75.4 \pm 0.5$ & $67.9 \pm 1.1$ \\ %\hline
 \textit{SP (via seg. masks)}     & $37.9 \pm 0.8$  & $39.8 \pm 1.3$ & $84.6 \pm 0.3$ & $4.5 \pm 0.2$ & $41.3 \pm 0.8$ & $77.2 \pm 0.9$ & $67.4 \pm 0.3$ \\ 
 \textit{SP (via annotators)} & $\textbf{41.6} \pm \textbf{1.3}$  & $46.2 \pm 0.7$ & $\textbf{85.7} \pm \textbf{1.3}$ & $\textbf{26.6} \pm \textbf{2.1}$ & $\textbf{44.3} \pm \textbf{0.8}$ & $\textbf{75.6} \pm \textbf{0.3}$ & $\textbf{68.7} \pm \textbf{0.4}$ \\ 
\hline
\end{tabular}
}
\label{table:hand_annotation}
\end{table*}

%------------------
\subsection{Further Comparison and Analysis}
\label{section:further-ana}
%------------------
For demonstration purpose, the SP information used in the previous experiments is simply obtained from the given annotated ground-truth segmentation masks. Certainly, in practice, we need the estimated SP information directly from annotators rather than from the ground-truth segmentation masks and thus to significantly simplify the data annotation process. Below we showcase that rough estimated SP directly from annotators can indeed be obtained efficiently and cheaply and is quite sufficient for our models to achieve good performance.

To directly obtain the SP annotations (in the absence of ground-truth masks), 52 images were randomly picked from the \texttt{Aerial Dubai} dataset, and then three annotators were asked to estimate the SP for the provided images. The estimated SP scores were then averaged. Afterwards, data augmentation techniques such as flipping and rotation were applied to obtain 416 images for training. 
Further details of the annotation process are given in Appendix.
Table \ref{table:annotation_efficiency} highlights the time and memory cost to produce the SP annotations compared to producing the ground-truth segmentation masks. 
Pixel annotation for a single image with 5 objects takes roughly 330 seconds which is around 16 times more than the time required for SP annotation\footnote{Average time taken for per-pixel annotation is estimated based on \cite{bearman2016s}.}. Regarding memory, a mask with the size of $224\times224$ takes up around 148 kB. With compression, this value can drop to as low as 4 kB, which is still roughly 200 times larger than the SP which consists of only 5 numbers. This huge efficiency brought by our proposed SP strategy is quite significant particularly for big datasets which are required for semantic segmentation.

We now further compare the semantic segmentation performance between the benchmark model with ground-truth segmentation maps and our SPSS with the SP simply obtained from the ground-truth segmentation maps and the rough SP produced by the annotators (the details of the annotation process are given in Appendix), separately.
Table \ref{table:hand_annotation} presents the results on the same test set used in Table \ref{table:Aerial_table}. The results are quite impressive as SPSS with the rough SP estimations surpasses not only the way of using the SP obtained by the ground-truth maps but also the benchmark model trained using the costly ground-truth maps.

%---------------------
\section{Discussion and Limitation}
\label{discussion}%
%---------------------
\textit{SP (semantic proportions)} for each training image is required as annotation/label information for the presented semantic segmentation model. In this work, we obtained these proportions from both the segmentation maps available for the chosen datasets and three annotators directly to demonstrate the effectiveness and robustness of our proposed SP-based methodology. We would like to stress that the reason why we benefited from the existing segmentation maps, {which seems controversial to our main aim at first glance}, is to show that the proposed methodology is feasible in the presence of SP. Arguably, reasonable proportions can be simply extracted from the ground-truth segmentation maps if they are annotated properly. Therefore, obtaining SP from the readily available maps to achieve our aim is sensible. Clearly, our goal is to train our proposed model when the segmentation maps are unavailable. It is evident from our experiments that obtaining SP annotation could be much cheaper than obtaining the precise segmentation maps particularly for data volumes in high dimensions. 
There are obviously various ways to obtain SP readily in the absence of the segmentation maps, such as by employing mechanical turks.  There may exist applications such as estimating the density of housing in a particular area where information may be extracted from other studies or even obtained from  pre-trained large language models, e.g., ChatGPT \cite{openai2022chatgpt}. 

The results that we present in Section \ref{sec:experiments} are promising and one may wonder if the exact proportions are a must, which would make the proposed setting as expensive as the traditional one. To demonstrate that it is not the case and that our methodology only needs rough SP, we presented sensitivity analysis regarding SP, where we added various amounts of noise to the extracted SP and demonstrated that the model performs satisfactorily well when trained with noisy SP. We also presented sensitivity analysis through investigating degraded SP by clustering to further support the robustness of our methodology when the precise SP is unavailable. The analysis suggests that our methodology not only works well with rough SP, but also with rough SP for only some representative images from the whole training set, indicating its need of significantly less annotation effort.     

\textit{Additional annotations}. 
In many scenarios, different types of annotations may exist. This raises the question that whether it is feasible for semantic segmentation methods to use the combination of different types of annotations to boost their performance. In this regard, our proposed semantic segmentation methodology based on SP delivers quite promising results. 

For datasets where the ground-truth segmentation maps are available, the SP annotation can be calculated directly. In these cases an additional loss function using the SP scores can be used as demonstrated by the SPSS+ model we have proposed. The results shown in Tables \ref{table:Aerial_table} and \ref{table:medical} demonstrated the good performance of SPSS+.  
The enhanced performance of our method  by utilising both annotation types may benefit from our introduced loss function $\mathcal{L}_{\rm total}$ in Eq. \eqref{eqn:loss-total}. It contains the  $\mathcal{L}_{\rm sp}$ loss defined in Eq. \eqref{eqn:loss-sp}, which measures the MSE between the predicted SP and the given SP. The visualisation results in Figure \ref{fig:further} showed that our  $\mathcal{L}_{\rm sp}$ loss may produce better segmentation than the loss directly measuring the segmentation maps (that the benchmark method uses) in terms of the binarisation ability. Therefore, combining the $\mathcal{L}_{\rm sp}$ loss with the $\mathcal{L}_{\rm sm}$ loss and then forming the $\mathcal{L}_{\rm total}$ loss could boost the semantic segmentation performance, e.g. see the visualisation given in Figure \ref{fig:mic_better}.

\textit{Limitations}. SP provides much less information than standard segmentation annotations.  In some scenarios, for example, with large number of classes or where some classes represent only a tiny proportion of any image, the semantic proportions might not provide enough information for the network to infer the classes.  Thus the utility of SP will be problem dependent.  In many ways the surprising observation for us was to discover how powerful SP is on a range of problems given how little information we are providing to the network.  Although SP will not be a solution for all segmentation problems, we believe that its relative cheapness means that it may be the method of choice in a number of applications where semantic segmentation is required, but the resources to hand annotation images is limited.

In this work, we proposed a new semantic segmentation methodology by introducing the SP annotation. In the scenario of quite limited annotation, using SP for semantic segmentation can already achieve competitive results. If additional annotations are available, our method can easily utilise them for performance boost. Moreover, for existing segmentation methods that use different types of annotations, we also suggest involving SP in these methods; e.g., our proposed $\mathcal{L}_{\rm sp}$ loss could be served as a type of regularisation given its effectiveness in binarisation.

%---------------------
%---------------------
\section{Conclusion}
\label{conclusion}
%---------------------
Semantic segmentation methodologies generally require costly annotations such as the ground-truth segmentation masks in order to achieve satisfying performance. Motivated by reducing the annotation time and cost for semantic segmentation, we in this paper presented a new methodology SPSS, relying on the SP annotation instead of the costly ground-truth segmentation maps. Extensive experiments validated the great potential of the proposed methodology in reducing the time and cost required for annotation, making it more feasible for large-scale applications. Furthermore, this innovative design opens up new opportunities for semantic segmentation tasks where obtaining the ground-truth segmentation maps may not be feasible or practical. We believe that the use of the SP annotation suggested in this paper offers a new and promising avenue for future research in the field of semantic segmentation, with evident and wide real-world applications.

\vspace{0.08in}

\textbf{Acknowledgements.} Halil Ibrahim Aysel is thankful for the support from the Republic of Turkiye Ministry of National Education.

%---------------------
%---------------------
\section*{Appendix}
\label{appendix}
%---------------------

%----
\begin{figure*}
    \centering
    \includegraphics[width=0.8\textwidth,height=0.5\textwidth]{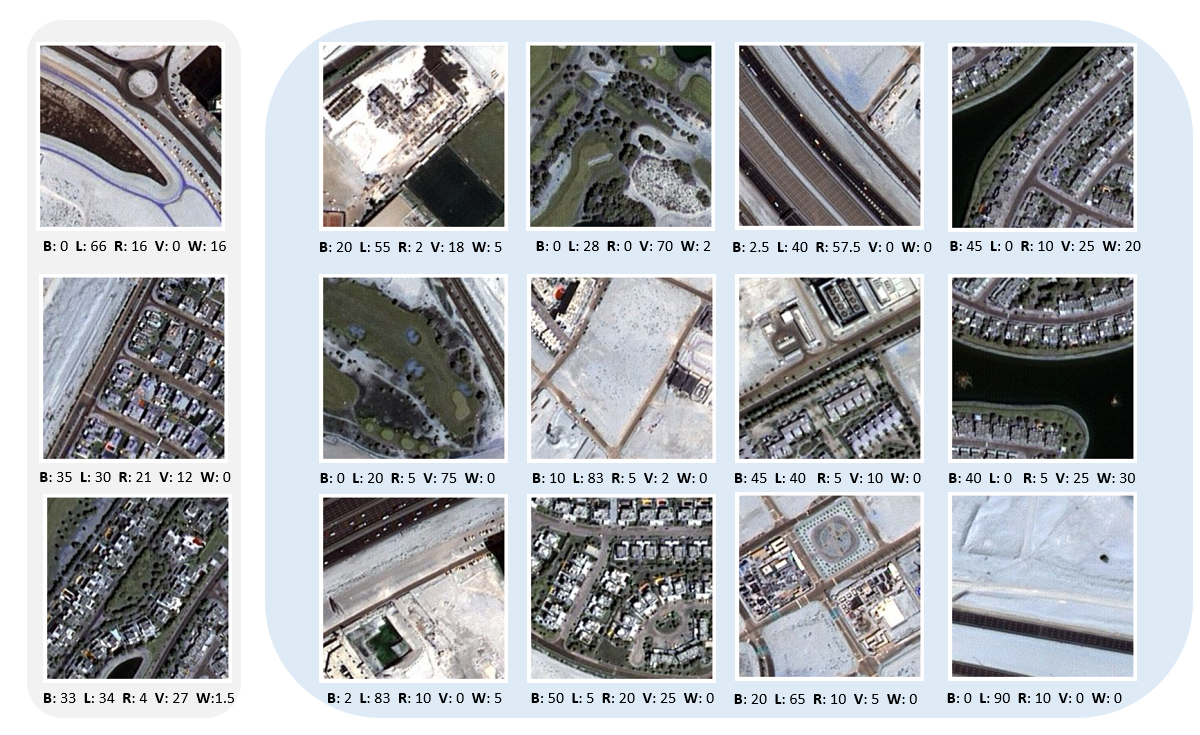}
    \put(-405, 255){\textbf{Reference Images}}
    \put(-243, 255){\textbf{Average SP Estimations by Annotators}}
     \put(-453, 157){ \textbf{Classes}}
    \put(-455, 144){\small \textbf{B:} building}
    \put(-455, 132){\small \textbf{L:} land}
    \put(-455, 120){\small \textbf{R:} road}
    \put(-455, 108) {\small \textbf{V:} vegetation}
    \put(-455, 96) {\small \textbf{W:} water}
    \caption{Showcase of the SP annotation process by annotators directly. Three annotators were asked to annotate a batch with $52$ images for training. Left: reference images whose SP information is calculated from the pixel-wise annotated ground-truth segmentation maps. Right: some randomly selected images with their average SP estimations by the three annotators.}
    \label{fig:hand_annotation}
\end{figure*}

\subsection{Semantic Proportions Annotation}
%--------

For the experiments presented in Section \ref{section:further-ana}, three annotators were asked to annotate a small batch containing $52$ images from the {\tt Aerial Dubai} dataset each with the size of $288\times288$ to show the efficiency of the SP annotation process compared to the pixel-wise annotation, as well as the excellent semantic segmentation ability of the proposed  SPSS model compared to the benchmark model (with the ground-truth segmentation maps).

\begin{itemize} [leftmargin=*]

\item The annotators were provided with three reference images whose SP information is simply obtained via the pixel-wise segmentation maps, see the left of Figure \ref{fig:hand_annotation} above. The reference images could be helpful for annotators to adjust their estimations; for instance, for the last image in the first row of Figure \ref{fig:hand_annotation} regarding the SP estimations, it is clear that the water area is a little larger than that in the first reference image, which helps the annotators to estimate a proportion with a larger value than that for the water area in the reference image (i.e., 20\% vs. 16\%). The average estimation of the three annotators for the water area in the mentioned image is around $20\%$, which is quite close to the value obtained by its ground-truth map, i.e., $21.3\%$, showing the efficiency of the SP annotation directly by annotators in this manner. Moreover, our sensitivity experiments showed that obtaining precise SP information for training is not a must for our SPSS model to perform well, making the SP annotation process even more efficient and relaxing given its tolerance of rough deviation in the SP estimations. 

\item After each annotator completed their SP annotation, the average SP annotation of the three annotators is obtained for the $52$ images. 

\item Finally, two types of augmentation strategies were carried out to increase the training dataset size. Each image was flipped horizontally and rotated by $90$, $180$ and $270$ degrees clockwise. The rotations were also applied to every flipped image. Therefore, $8$ images were obtained for every image, and a training dataset consisting of $416$ images in total is formed. Note that, since the SP information is irrelevant to the position of the content in an image, the estimated SP for one image is also applied to all of its 7 augmented versions.

\end{itemize}

\subsection{Algorithm}

Algorithm  \ref{alg:algorithm-1} shows the noise injection processes for the experiments presented in Section \ref{subsec:sp-de-noise}.

\begin{algorithm*}[ht]
  \caption{Noisy SP $\vect{\tilde{\rho}}^*_{i}$ Generation}
  \label{alg:algorithm-1}
  \begin{algorithmic}[1]
   \State \textbf{Input:}
    Ground-truth SP $\vect{\rho}^*_{i}$ of image $\vect{X}_i$, standard deviation $\sigma$, lower bound $a$, and upper bound $b$.
   \State \textbf{Output:}
    Noisy SP $\vect{\tilde{\rho}}^*_{i}$ 
    \If {${\rm length}(\vect{\rho}^*_{i})$ == 1} \Comment{E.g., medical imaging datasets}
        \State Randomly select $\lambda$ from $\{-1, 1\}$; 
        \State $\tilde{\rho}^*_{i1} = \rho^*_{i1} + \lambda \ \mathcal{U}(a, b) \rho^*_{i1}$;
  \Else \Comment{E.g., {\tt Aerial Dubai} dataset}
    \For {$j = 1$ to {${\rm length}(\vect{\rho}^*_{i})$}}
    \State $
    \tilde{\rho}^*_{ij} = \rho^*_{ij} + \mathcal{N}(0, \sigma)$;
 \EndFor
 \EndIf

\State \textbf{return} $\vect{\tilde{\rho}}^*_{i}$ 

\end{algorithmic} \label{Alg:noisy-SP}
\end{algorithm*}

\bibliographystyle{IEEEtran}
\bibliography{main}

\end{document}